Research and Applications

# DiMB-RE: mining the scientific literature for diet-microbiome associations

**Gibong Hong, MS[1], Veronica Hindle, MS[2], Nadine M. Veasley, MS[3], Hannah D. Holscher, PhD[2,3,4], Halil Kilicoglu, PhD*,[1,3,4]**

[1]School of Information Sciences, University of Illinois Urbana-Champaign, Champaign, IL 61820, United States, [2]Department of Food Science and Human Nutrition, University of Illinois Urbana-Champaign, Urbana, IL 61801, United States, [3]Division of Nutritional Sciences, University of Illinois Urbana-Champaign, Urbana, IL 61801, United States, [4]Personalized Nutrition Initiative, University of Illinois Urbana-Champaign, Urbana, IL 61801, United States

*Corresponding author: Halil Kilicoglu, PhD, School of Information Sciences, University of Illinois Urbana-Champaign, 501 E Daniel Street, Champaign, IL 61820, United States (halil@illinois.edu)

## Abstract

**Objectives:** To develop a corpus annotated for diet-microbiome associations from the biomedical literature and train natural language processing (NLP) models to identify these associations, thereby improving the understanding of their role in health and disease, and supporting personalized nutrition strategies.

**Materials and Methods:** We constructed DiMB-RE, a comprehensive corpus annotated with 15 entity types (eg, Nutrient, Microorganism) and 13 relation types (eg, INCREASES, IMPROVES) capturing diet-microbiome associations. We fine-tuned and evaluated state-of-the-art NLP models for named entity, trigger, and relation extraction as well as factuality detection using DiMB-RE. In addition, we benchmarked 2 generative large language models (GPT-4o-mini and GPT-4o) on a subset of the dataset in zero- and one-shot settings.

**Results:** DiMB-RE consists of 14 450 entities and 4206 relationships from 165 publications (including 30 full-text Results sections). Fine-tuned NLP models performed reasonably well for named entity recognition (0.800 $F_1$ score), while end-to-end relation extraction performance was modest (0.445 $F_1$). The use of Results section annotations improved relation extraction. The impact of trigger detection was mixed. Generative models showed lower accuracy compared to fine-tuned models.

**Discussion:** To our knowledge, DiMB-RE is the largest and most diverse corpus focusing on diet-microbiome interactions. Natural language processing models fine-tuned on DiMB-RE exhibit lower performance compared to similar corpora, highlighting the complexity of information extraction in this domain. Misclassified entities, missed triggers, and cross-sentence relations are the major sources of relation extraction errors.

**Conclusion:** DiMB-RE can serve as a benchmark corpus for biomedical literature mining. DiMB-RE and the NLP models are available at https://github.com/ScienceNLP-Lab/DiMB-RE.

## Background and significance

Nutrition plays an essential role in maintaining health and preventing chronic diseases. It is increasingly recognized that human responses to dietary input are driven by unique host and microbiome features. This gave rise to the personalized nutrition paradigm,[1,2] which emphasizes the analysis of clinical, omics, behavioral, and environmental data using machine learning techniques to predict individualized responses to diet and tailor dietary interventions.[3] While such data-driven approaches can predict diet effects,[3] they often lack mechanistic interpretation, needed for broad acceptance of dietary interventions by practitioners and patients.[1,4]

Research literature offers valuable scientific evidence on diet and its effects on microbiome and human health.[5] However, much of the evidence remains buried in unstructured text, making it difficult to leverage it for computational analysis. The rapid growth in the literature exacerbates this problem. While some curated databases exist,[6–8] they are often limited (eg, microbe-disease associations) and require manual efforts, restricting their usefulness and scalability. Natural language processing (NLP) can help address these challenges by machine-reading literature to extract information on diet-microbiome associations and their relation to human health and disease. Structuring this information (eg, knowledge graphs) and combining it with patient data can facilitate knowledge-guided analyses and advance personalized nutrition.

There has been growing interest in using NLP for extracting information related to diet and/or microbiome from the literature in recent years,[9–22] most focusing on a few types of information (eg, bacteria entities[9–12]; microbe-disease,[14,15] food-disease associations,[18] or food-drug interactions[19]) using approaches ranging from dictionary-based methods[9,13] to fine-tuned BERT-based models[18,21,22] and few-shot learning with generative large language models (LLMs)[17] for named entity recognition (NER) and relation extraction (RE). Despite these advances, diverse and sizable corpora that consider a broader set of relevant associations that better serve the needs of personalized nutrition are lacking. Incorporating relevant domain vocabularies would increase the utility of such corpora for knowledge base construction and knowledge integration.[5,23,24]

In the broader context of biomedical NLP, many corpora have been developed to facilitate model development, although most of them also focus on specific entity types, such as diseases[25] or chemicals,[26] or a single relation type, such as protein-protein interactions[27] or drug adverse effects.[28] The recent BioRED corpus[29] is more comprehensive, including 8 relation types between 6 entity types, although most relation types are very rare. In contrast, biological event extraction corpora, which focus on molecular level events, often include a larger number of relation types[30,31] but a small number of entity types. The current standard approach for NER and RE from biomedical literature is to fine-tune domain-specific pretrained models, such as PubMedBERT (also known as BioMedBERT),[32] on annotated corpora. We refer the reader to Zhao et al[33] for an in-depth discussion of recent biomedical NER and RE methods.

In this work, we present DiMB-RE, a corpus of 165 nutrition and microbiome-related publications, aimed at addressing existing challenges in diet and microbiome information extraction. We also establish robust baseline information extraction models for this corpus. Specifically, we make the following contributions:

1) We annotated titles and abstracts of 165 publications with 15 entity types (eg, Nutrient, Microorganism) and 13 relation types that hold between them (eg, INCREASES, IMPROVES).
2) We annotated Results sections of 30 articles (out of 165) to assess the impact of the information from full text.
3) We annotated relation triggers and certainty information to ground and contextualize relations.
4) We trained and evaluated NER and RE models using the state-of-the-art pretrained language models as well as generative LLMs.

## Materials and methods

### Corpus construction

#### Data collection

Domain experts in food science and human nutrition manually created the PubMed search string to collect articles on nutrition and microbiome written in English for annotation. The search string terms were based on current diet-microbiome research, and included microbial genus and phyla, metabolites, metabolic health markers, health outcomes, and dietary approaches deemed important for personalized nutrition, and are provided in Appendix S1.

#### Annotation

Annotation was performed over several stages using Brat annotation tool.[34] Initially, 5 articles (titles and abstracts) were annotated by 2 food science and nutrition graduate students (V.H., N.M.V.) and the senior investigator (H.K.), an expert in NLP/biomedical informatics, to determine relevant entity and relation types and develop draft annotation guidelines. The data model and the annotation guidelines were refined in discussions with the senior investigator with domain expertise (H.D.H.). In the next stage, 20 articles were annotated by the graduate student annotators, interannotator agreement was calculated, and annotations were adjudicated. Annotators consulted with the senior investigators to resolve disagreements. Data model and annotation guidelines were refined. The annotation/adjudication process was repeated in 3 stages, on 30 and 15 abstracts, and 4 Results sections, respectively. Finally, the remaining articles were split between the 2 annotators, who individually annotated the articles (including the Results sections in some cases), in consultation with the senior investigators. H.K. and G.H. verified the final annotations for consistency and accuracy. The data model is presented in the section below. An annotation example is provided in Figure 1. The annotation guidelines are provided in Appendix S2.

#### Data model

The data model used for annotation consists of 15 entity types and 13 relation types. The entity types are Food, Nutrient, DietPattern, Microorganism, DiversityMetric, Metabolite, Physiology, Disease, Measurement, Enzyme, Gene, Chemical, Methodology, Population, and Biospecimen. The last 3 types were captured as study-level characteristics. Nested entities were allowed in annotation (eg, in the phrase *PC-enriched virgin olive oil*, *PC* was annotated as Nutrient and the entire phrase as Food; *PC* refers to phenolic compounds).

Thirteen relation types were adapted from the UMLS Semantic Network[35,36] and the Biolink model,[37] which formalizes entity and relationship types for translational science. These types are as follows: AFFECTS, IMPROVES, WORSENS, ASSOCIATED_WITH, POS_ASSOCIATED_WITH, NEG_ASSOCIATED_WITH, INTERACTS_WITH, INCREASES, DECREASES, CAUSES, PREVENTS, PREDISPOSES, and HAS_COMPONENT. Most relation types are directional with an explicit Agent and Theme argument, with the exceptions of ASSOCIATED_WITH and INTERACTS_WITH, which have 2 Theme arguments and are used when no directionality can be inferred from text. Relation annotations also include triggers that provide justification for the identified relationships (eg, *ameliorates* for IMPROVES).

We adopted the 6-way factuality categorization from prior work[38] to characterize the certainty levels that the authors associate with the relationship statements. These categories are Factual, Probable, Possible, Doubtful, Negated, and Unknown, capturing certainty on an ordinal scale.

Definitions, examples, and other details for entity and relation types are provided in the annotation guidelines (Appendix S2).

### Interannotator agreement

We assessed interannotator agreement (IAA) on 65 abstracts and 4 Results sections that were double-annotated to ensure the consistency of the corpus. In line with previous work,[39] to calculate IAA, we used $F_1$ score when taking 1 set of annotations as predictions and the other annotator's labels as ground truth. We calculated 2 scores: IAA (Exact) follows strict boundary matching for entity annotations, while IAA (Partial) requires type match while allowing overlapping entity spans. This distinction helps us quantify annotation challenges due to entity boundaries. When calculating IAA for relations, the same principle is applied to the entity mentions involved in the relation.

### NLP models

Using DiMB-RE, we conducted experiments in NER and RE by fine-tuning BiomedBERT,[32] a state-of-the-art, domain-specific language model trained on PubMed articles, as our base model. Additionally, we investigated whether trigger annotations are useful for RE, and explored in-context learning with generative LLMs in zero- and one-shot settings. While DiMB-RE includes cross-sentence relations, we limited our experiments to sentence-bound relations, as cross-sentence relations are significantly more challenging. Due to lower IAA on Results sections, we used annotations from this section for training only. We provide the experimental settings and their environmental impacts in Appendix S5.

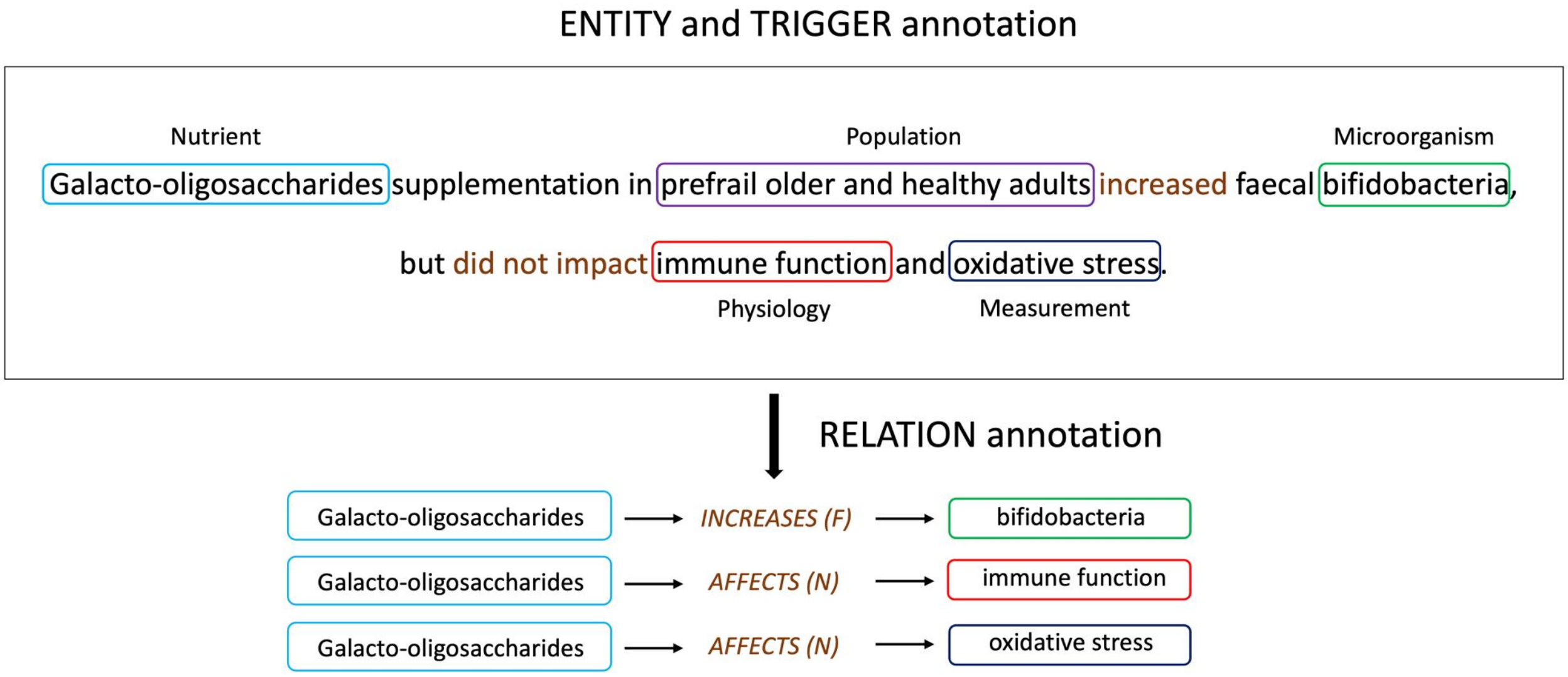

**Figure 1.** Annotation of a sentence from a PubMed abstract (PMID: 33509667). Entity mentions are shown in colored boxes and triggers and the corresponding relations are highlighted in brown. Relations are assigned certainty levels. Abbreviations: F, Factual; N, Negated.

### Fine-tuning approach

We used a pipeline approach for RE, where NER is followed by relation classification, which predicts whether each entity pair is related and if so, the type of the relation. We employed 2 state-of-the-art architectures, PL-Marker[40] and PURE,[41] both of which use 2 independent encoder-only models for NER and RE. We describe the better-performing PL-Marker architecture below (Figure 2) and the PURE model in Appendix S3.

We formulate NER as span prediction.[42] While sequence labeling has been the more conventional setup for NER,[43] we used span prediction because it handles nested entities more naturally. In span prediction, all possible spans up to a predetermined length (eg, 5 tokens) are enumerated as candidate entity spans and classified into entity types (or None).[42] PL-Marker[40] uses *levitated markers* to model the interrelation between entity spans during the encoding phase. Each levitated marker pair includes start and end markers for a span, sharing position embeddings with the span's tokens while preserving the original token order. Directional attention ensures that each marker interacts only with its pair and text tokens, enabling efficient parallel modeling of multiple spans in a single sequence. To prevent marker overflow in the input, PL-Marker NER model employs a *neighborhood-oriented packing* strategy, which organizes levitated markers into batches for improved efficiency. Specifically, spans with the same start token are grouped together, allowing the model to compare adjacent spans and better distinguish entity boundaries.

PL-Marker RE model uses span pair classification for entities extracted by the NER model. It adopts a *subject-oriented packing* strategy, where solid markers[44] are inserted around the subject span to highlight its role, while levitated markers are used for candidate object spans. To obtain the representation of span pairs, the RE model concatenates the contextualized representation of start and end solid markers and levitated markers then passes this representation through a feedforward network for relation classification. This subject-oriented packing strategy enables the model to focus on the relationships involving the same subject span while reducing time complexity. By integrating entity and relation modeling, PL-Marker achieves an advantage over methods that process span pairs sequentially.[41] We use cross-entropy loss to fine-tune both the NER and RE models.

### Enhancing RE with triggers

In DiMB-RE, we annotated relation triggers and linked them to relation arguments, which is not common in recent relation corpora, but was sometimes done in event extraction corpora.[45] To examine whether the trigger spans could contribute to RE, we conducted experiments by incorporating trigger information into the RE model. We recognized triggers in a similar manner to entities, using span representations and classifying the trigger type (corresponding to a relation type). We set the same maximum span length for entity and trigger candidate spans and used a single feedforward layer to recognize them. To incorporate trigger information for RE, we extended the *subject-oriented packing* strategy by pairing each subject with a trigger (*subject-trigger-oriented packing*). This pair acts as a key for selecting object spans. For example, in a sentence with 3 entities and 1 trigger, each entity can serve as a subject, with the others as objects, forming triples like [,;<TRG>;</TRG>;<OBJ1>;</OBJ1>] and [,;<TRG>;</TRG>;<OBJ2>;</OBJ2>]. Relation model is trained to determine whether the given triple represents a valid relationship.

### Factuality detection

Once the relations are identified, we assign them certainty levels, which we formulate as multiclass classification. Although we annotated 6 certainty levels, we simplified them into Factual, Negated, and Uncertain (combining Probably, Possible, Doubtful, and Unknown) due to limited examples with these types in DiMB-RE. This 3-way classification also aligns better with Certainty and Negation dimensions used in event extraction corpora.[45] This task uses the same input representation as that for RE (Figure 2).

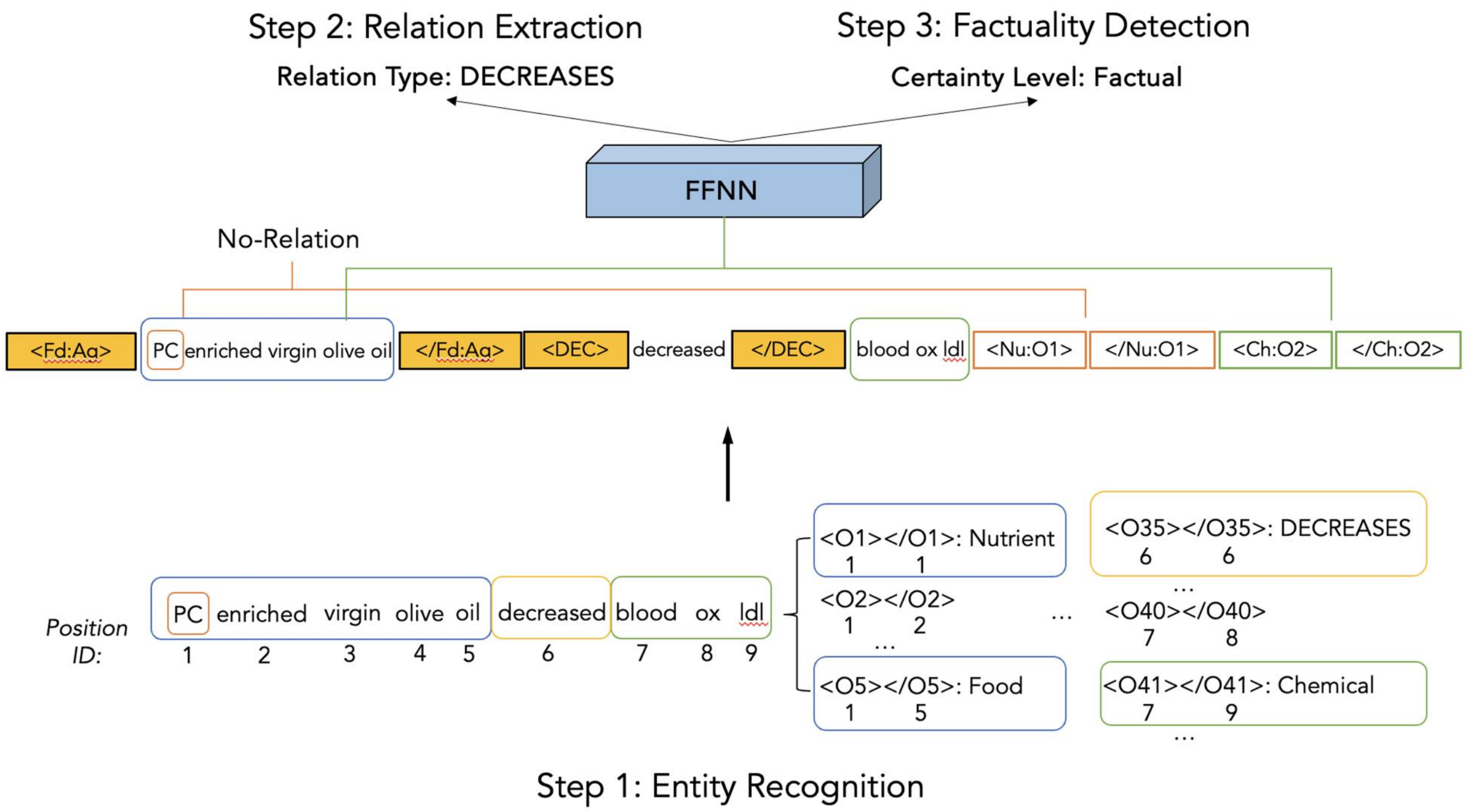

**Figure 2.** Overview of our model architecture based on PL-Marker architecture.[40] We take a 3-step pipeline approach: (1) entity and trigger recognition (named entity recognition—NER), (2) relation extraction (RE), and (3) factuality detection. Input to the model are individual sentences of the document. The model predicts entity mentions (with types) and their relations as well as the certainty levels of the predicted relations. In this example, the predicted relation is *PC enriched virgin olive oil*-DECREASES-*blood ox-ldl*, with the certainty level Factual.

### In-context learning with LLMs

As in other NLP tasks, LLMs are increasingly applied to biomedical information extraction.[46–48] While zero-/few-shot information extraction with LLMs often underperforms fine-tuned models,[46,49] their capabilities are improving. In this work, we used ground truth entities as the basis for zero- and one-shot RE with GPT-4o-mini (*gpt-4o-mini-2024-07-18*) and GPT-4o (*gpt-4o-2024-08-06*),[47] the most advanced OpenAI models in the small and large models categories, respectively, at the time of our experiments. We adapted an RE prompt[46] and applied task-aware retrieval.[50] We provide our prompt template along with an example in Appendix S4. Large language model experiments were carried out on October 21, 2024.

The prompt template includes a Task Description, Guideline, demonstration, and test sample. The Task Description explains the task specifics, while the Guideline outlines relation types with definitions, examples, and the rationale for the relation in the example. This is followed by a demonstration selected through task-aware retrieval,[50] which uses kNN retrieval based on SimCSE[51] to find the training examples most similar to the test sample. The RE prompt has 3 multiple-choice questions: (1) whether a relation exists (Relation/No relation), (2) possible relation types (filtered to those occurring at least 3 times between the entity types in the training set), and (3) the certainty level. Special markers highlight the ground-truth entities in the input samples.

### Evaluation

For IE evaluation, we use the standard evaluation metrics; precision, recall, and their harmonic mean, $F_1$ score. We report strict and relaxed metrics for both NER and RE. In relaxed evaluation, a predicted named entity counts as a true positive if its span overlaps with the ground truth span and both share the same entity type. In strict evaluation, the span boundaries must match, as well. For RE, in both cases, relation types must match. Although we use trigger information for RE, we do not require trigger matching in RE, as we identify triggers mainly to improve RE. We also use precision, recall, and $F_1$ for factuality detection evaluation and limit it to correctly identified relations only, because those for incorrect relations will always be invalid. To estimate the end-to-end performance, we also consider the certainty level of the predicted relations. We calculate 95% confidence intervals (CIs) for all evaluation metrics using bootstrap sampling.

**Table 1.** Descriptive statistics of DiMB-RE.

| | Title-Abstract (median) | Results (median) | Total |
|---|---|---|---|
| Documents | 165 | 30 | 165 |
| Sentences | 2181(13) | 2129 (28.5) | 4310 |
| Words | 68 136 (364) | 68 911 (944.5) | 130 787 |
| Entities | 7776 (46) | 6674 (196.5) | 14 450 |
| Triggers | 1536 (9) | 701 (18.5) | 2237 |
| Relations | 2831(15) | 1375 (41.5) | 4206 |

Total number and median (number in parentheses) are presented for each element.

## Results

### Corpus statistics

We provide the descriptive statistics of DiMB-RE in Table 1. On average, Title-Abstract sections are shorter than Results, contain a smaller number of entities but more triggers and relations when normalized by length.

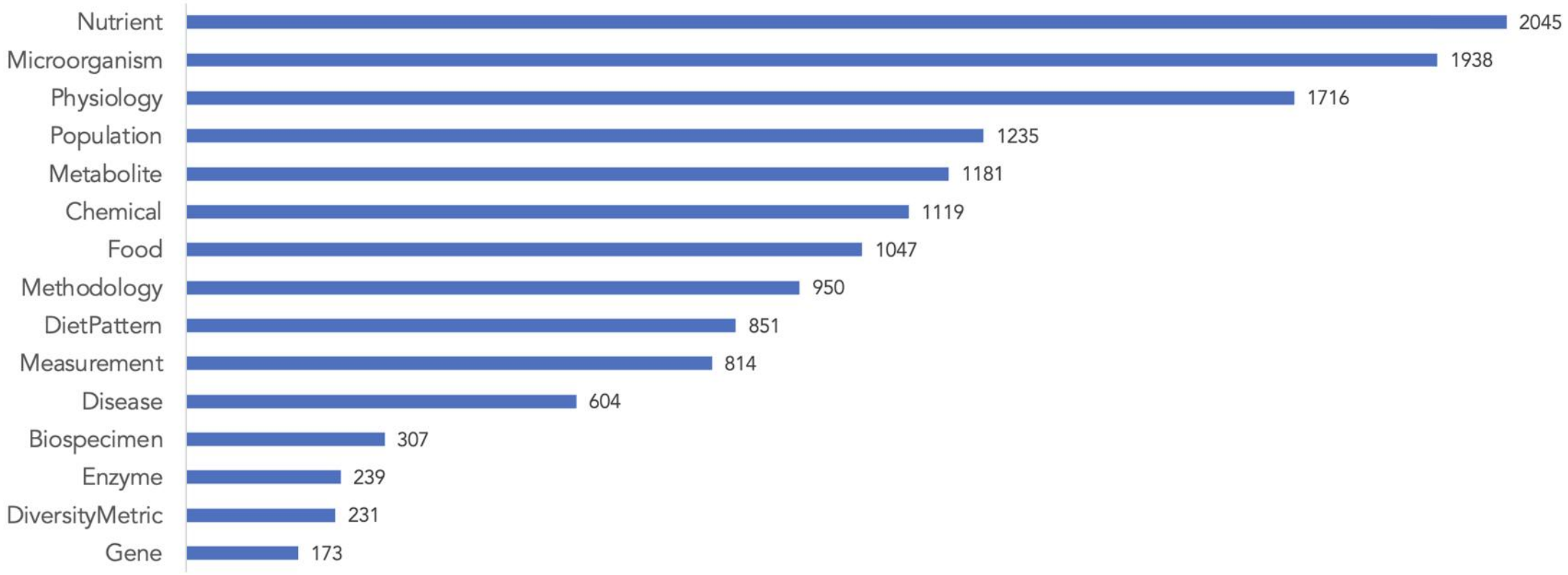

**Figure 3.** Label distribution of entity instances in DiMB-RE.

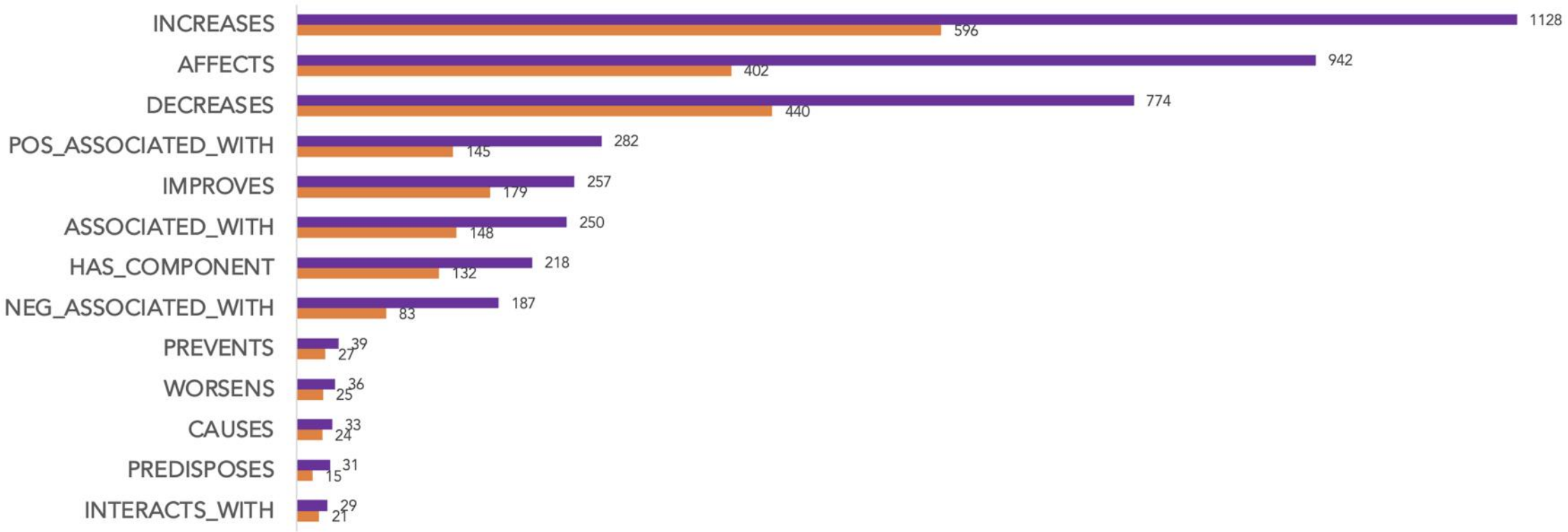

**Figure 4.** Class-level trigger and relation type distributions in DiMB-RE.

Figure 3 presents the label distribution of entities. The most common entity types are Nutrient (14.2%), Microorganism (13.4%), and Physiology (11.9%), whereas Gene is the least frequent (1.2%). Around 2% of annotated entity mentions contain discontinuous spans. Around 5% of the annotated entity mentions are nested, with maximum depth of 2.

Figure 4 shows the class-level trigger and relation distribution. The most common relation types are INCREASES (26.8%), followed by AFFECTS (22.4%) and DECREASES (18.4%). AFFECTS has the highest relation/trigger ratio (2.34). Five relation types (CAUSES, INTERACTS_WITH, PREDISPOSES, PREVENTS, and WORSENS) each account for less than 1% of all relations. Approximately 11% of relations in the Title-Abstract subset and 13% of relations in the Result subset are cross-sentence relations.

Figure 5 provides the distribution of certainty levels of relations in the corpus. The majority of the relations are asserted as factual. Among relations stated with uncertainty, most were labeled as Unknown (74%), followed by Possible (19%), Probable (6%), and Doubtful (1%).

## IAA results

Over 4 phases, IAA (Exact) ranged from 0.66 to 0.76 for entity mentions (mean: 0.69) and 0.27 to 0.52 for relations (mean: 0.41). With IAA (Partial), these figures increased to 0.77-0.87 for entities (mean: 0.80) and to 0.44-0.66 for relations (mean: 0.54). For relations, matching of both arguments and triggers was required. The overall relation agreement increases from 0.41 to 0.44, if no trigger matching is required. Agreement on Results section annotations was lower than on abstracts, particularly for relations. Agreement on abstract annotations improved over time.

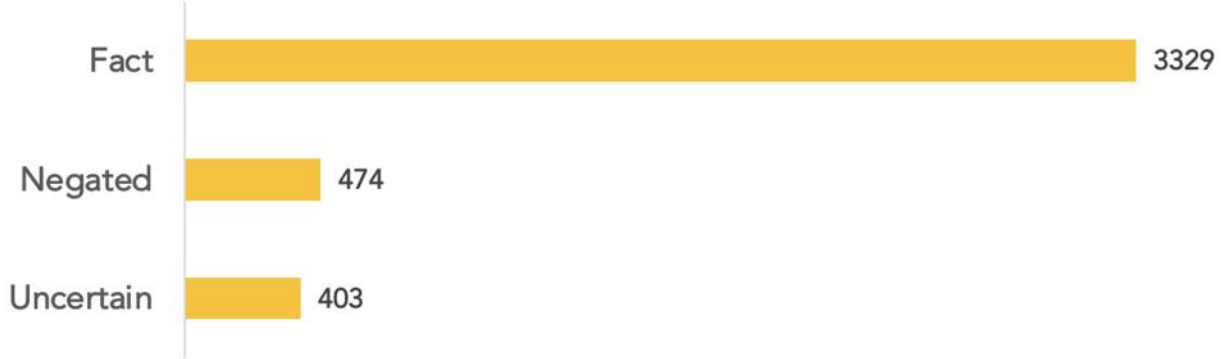

**Figure 5.** Distribution of certainty labels in DiMB-RE.

## NLP model performance

The performances of the PL-Marker-based models are presented in Table 2. PURE performance and its comparison to PL-Marker performance are provided in Appendix S6. The training data for both models includes Titles, Abstracts, and Results sections.

Compared to strict evaluation, $F_1$ scores for both NER and trigger detection increased by 6.2 points in relaxed evaluation. For individual entity types, the performance ranges from 0.934 for Enzyme to 0.611 for DiversityMetric. Among trigger types, we obtained the highest $F_1$ score for WORSENS (0.893) and lowest for PREVENTS (0.293).

The RE model yields 0.458 $F_1$ score in strict evaluation (0.519 in relaxed). The end-to-end model yields 0.445 $F_1$ score with strict evaluation. We consider this as our main

**Table 2.** PL-Marker model performance on the test set.

| | | Precision [95% CI] | Recall [95% CI] | $F_1$ [95% CI] |
|---|---|---|---|---|
| NER | Strict | 0.787 [0.778-0.797] | 0.814 [0.806-0.820] | 0.800 [0.793-0.808] |
| | Relaxed | 0.861 [0.852-0.870] | 0.863 [0.857-0.869] | 0.862 [0.855-0.869] |
| Trigger detection | Strict | 0.652 [0.640-0.664] | 0.691 [0.682-0.698] | 0.671 [0.662-0.678] |
| | Relaxed | 0.712 [0.703-0.723] | 0.755 [0.743-0.766] | 0.733 [0.724-0.738] |
| RE | Strict | 0.463 [0.452-0.473] | 0.454 [0.435-0.471] | 0.458 [0.445-0.472] |
| | Relaxed | 0.524 [0.514-0.533] | 0.515 [0.488-0.542] | 0.519 [0.502-0.535] |
| Factuality detection | | 0.965 [0.959-0.970] | 0.524 [0.512-0.534] | 0.679 [0.670-0.686] |
| End-to-end | Strict | 0.449 [0.438-0.463] | 0.441 [0.417-0.458] | 0.445 [0.432-0.459] |
| | Relaxed | 0.508 [0.499-0.518] | 0.499 [0.473-0.521] | 0.503 [0.488-0.518] |

We provide mean performances over 5 training runs with different seeds, along with 95% CIs based on bootstrap sampling (in square brackets).
Abbreviations: CI, confidence interval; NER, named entity recognition; RE, relation extraction.

**Table 3.** PL-Marker model's RE and end-to-end pipeline performance with gold entities and triggers.

| | Precision [95% CI] | Recall [95% CI] | $F_1$ [95% CI] |
|---|---|---|---|
| RE (gold entities) | 0.735 [0.726-0.748] | 0.663 [0.657-0.669] | 0.697 [0.690-0.703] |
| RE (gold entities and triggers) | 0.815 [0.809-0.825] | 0.799 [0.792-0.806] | 0.807 [0.800-0.813] |
| End-to-end (gold entities) | 0.697 [0.683-0.712] | 0.629 [0.623-0.635] | 0.661 [0.653-0.669] |
| End-to-end (gold entities and triggers) | 0.773 [0.767-0.782] | 0.757 [0.752-0.762] | 0.765 [0.760-0.771] |
| *Generative models* | | | |
| *Zero-shot (with gold entities)* | | | |
| GPT-4o-mini RE | 0.275 | 0.396 | 0.325 |
| GPT-4o RE | 0.274 | 0.604 | 0.377 |
| GPT-4o-mini end-to-end | 0.232 | 0.333 | 0.274 |
| GPT-4o end-to-end | 0.255 | 0.563 | 0.351 |
| *One-shot (with gold entities)* | | | |
| GPT-4o-mini RE | 0.355 | 0.458 | 0.400 |
| GPT-4o RE | 0.305 | 0.604 | 0.406 |
| GPT-4o-mini end-to-end | 0.323 | 0.417 | 0.364 |
| GPT-4o end-to-end | 0.284 | 0.563 | 0.378 |

Mean performance results over 5 training runs with different seeds, along with 95% confidence intervals based on bootstrap sampling, are provided (in square brackets). Note that the generative model evaluation is limited to 10% of the test set.
Abbreviations: CI, confidence interval; RE, relation extraction.

end-to-end model. Based on end-to-end performance, the highest performing relation types are INTERACTS_WITH, INCREASES, and DECREASES and the lowest is ASSOCIATED_WITH. Class-level NER, trigger detection, and RE performance results are provided in Appendix S7.

We examined the effect of using gold entity and trigger mentions in RE (Table 3). Using GOLD entities increased RE performance by 23.9 percentage points (0.458-0.697 $F_1$). Adding GOLD triggers further improves the performance by 11 percentage points (0.697-0.807 $F_1$). Using gold entities and triggers increases the end-to-end performance significantly, as well (20.3 and 30.7 points, respectively).

We evaluated GPT-based inference using gold entities (Table 3). GPT-4o performed better than GPT-4o-mini in zero-shot setting, while their performances were similar in one-shot setting. However, the performance of both models significantly lagged supervised RE methods, consistent with previous studies.[50,52] Both GPT models significantly overpredicted labels for the entity pairs that are unrelated. Due to computational costs, we only processed 10% of the test set with GPT models, so these results must be taken with caution.

We conducted an ablation analysis to understand whether using Results section annotations, despite lower IAA, helps with NLP models. The results show that the models benefit from using Results section annotations (Table 4).

## Discussion

### Corpus

We have constructed DiMB-RE, a corpus of scientific publications with more than 14K entities and 4K relations on diet-microbiome information, which, to our knowledge, is the largest and most diverse dataset focusing specifically on this domain. In terms of diversity of entity and relation types, it also compares favorably to other biomedical literature corpora, such as BioRED.[29] While DiMB-RE contains fewer articles, the number of entities and relations is similar to that in the BioRED training set (400 abstracts), indicating that the documents in our dataset are more information-dense. Unlike BioRED, we represent relation directionality when relevant in our annotation and include relation triggers, generally ignored in biomedical relation corpora.

Interannotator agreement was reasonable for entities, but modest for relations. Agreement in Results sections was lower, suggesting the increased difficulty of annotating full-text articles. Based on the lower agreement on Results sections, we used them for training only. Among entity types, the highest agreement was obtained for Microorganism and lowest for DietPattern. The former is often clearly delineated in text, while the latter can consist of more complex phrases as well as other dietary terms (ie, Food, Nutrient), which led to more disagreement. Among relation types, INCREASES had the highest agreement and INTERACTS_WITH the lowest. All

**Table 4.** Ablation analysis of the effect of using Results section annotations with the PL-Marker model with trigger information on the test set.

| | Precision [95% CI] | Recall [95% CI] | $F_1$ [95% CI] |
|---|---|---|---|
| NER | 0.787 [0.778-0.797] | 0.814 [0.806-0.820] | 0.800 [0.793-0.808] |
| NER w/o Results | 0.759 [0.754-0.763] | 0.789 [0.784-0.795] | 0.774 [0.770-0.777] |
| Trigger detection | 0.652 [0.640-0.664] | 0.691 [0.682-0.698] | 0.671 [0.662-0.678] |
| Trigger detection w/o Results | 0.649 [0.631-0.667] | 0.644 [0.636-0.650] | 0.648 [0.638-0.657] |
| RE (gold entities and triggers) | 0.815 [0.809-0.825] | 0.799 [0.792-0.806] | 0.807 [0.800-0.813] |
| RE (gold entities and triggers) w/o Results | 0.813 [0.805-0.819] | 0.757 [0.748-0.765] | 0.784 [0.776-0.791] |
| End-to-end (gold entities and triggers) | 0.773 [0.767-0.782] | 0.757 [0.752-0.762] | 0.765 [0.760-0.771] |
| End-to-end (gold entities and triggers) w/o Results | 0.769 [0.764-0.773] | 0.716 [0.710-0.723] | 0.741 [0.737-0.746] |

We provide mean performances over 5 training runs with different seeds, along with 95% CIs based on bootstrap sampling (in square brackets).
Abbreviations: NER, named entity recognition; RE, relation extraction.

disagreements were resolved by the annotators and the remaining annotations were verified by the senior investigator, which increases our confidence in the quality of the final corpus. We believe DiMB-RE can serve as a valuable resource for advancing the understanding of the complex interactions between diet, human metabolism, and gut microbiota.

## NLP models

We used the state-of-the-art PL-Marker model as the backbone of our experiments. Named entity recognition performance is reasonably high (0.800 strict $F_1$), but falls at the lower end of the performance range reported for PL-Marker and other state-of-the-art models on biomedical NER benchmarks (0.80-0.94).[53–55] This suggests that DiMB-RE is more challenging than similar benchmarks, probably because it includes more entity types and those are commonly less studied. Relaxed evaluation yields 6.2 points higher $F_1$, indicating that boundary detection remains a problem. The performance for trigger detection is lower than NER (0.671 vs 0.800 $F_1$), possibly because triggers are more heterogeneous in terms of relation types that they indicate. Without using Results section annotations, the performance was lower (2.6 points for entities and 2.3 for triggers), suggesting that even with some potential noise, additional data is useful for improving performance. Relation extraction results were overall modest (0.458 strict, $F_1$, 0.519 relaxed). Using gold entities yields 0.697 $F_1$, which is significantly lower than the state-of-the-art models in most biomedical RE benchmarks (0.80-0.85 $F_1$),[54] which, again, suggests that DiMB-RE is more challenging, likely because of the number of relation types considered and the imbalanced dataset. Using gold entities and triggers yields 0.807 $F_1$, indicating that improving NER and trigger detection performance further would have a significant impact on the performance. Some of the infrequent relation types on which the model performed poorly have been annotated in other corpora,[26] and incorporating annotations from those datasets might improve performance.[56]

Owing to its more advanced span representation and efficient packing strategies, PL-Marker outperforms PURE in all tasks, performance difference ranging from 1.1 points in trigger detection to 11 points in RE and end-to-end performance (relaxed).

The effect of trigger detection on RE and end-to-end performance was mixed. While using ground truth triggers clearly improved results for PL-Marker, using predicted triggers did not yield an improvement. The relatively strong RE performance of PL-Marker may have overshadowed the potential additional benefits of trigger detection. It is also possible that further refinements to subject-trigger-oriented packing is needed to observe benefits from trigger detection. On the other hand, trigger detection overall improved the RE and end-to-end performance of PURE, although this improvement was not uniform across relation types.

As end-to-end performance, we also consider whether the extracted relations are assigned the correct certainty labels. This yields 0.445 $F_1$ score (strict), which we consider our main result. Using ground truth entities, this increases to 0.661 and, with the addition of ground truth triggers to 0.765, indicating the important role of NER and trigger detection. We present an analysis of the PL-Marker errors in Appendix S8.

Using GPT-based models in zero- and one-shot setting yielded modest results, mainly due to overpredicting relations (low precision), which is in line with prior findings.[49,50] The performance difference between the models was higher in zero-shot setting, indicating that GPT-4o encodes more implicit knowledge about the task. Around 90% of all candidate entity pairs in our test set are NULL instances. Limiting the evaluation to entity pairs that are known to hold some relation, GPT-4o-mini yields 0.629 $F_1$ for RE (0.589 end-to-end), and GPT-4o yields 0.753 $F_1$ for RE (0.720 end-to-end). These results indicate that successfully moderating the tendency of generative models to classify NULL cases to valid labels is crucial for few-shot RE. Among the models, PL-Marker has a significantly lower environmental impact than PURE and GPT-4o (Appendix S5).

## Limitations

Our study has limitations. First, about 11% of relations in the Title-Abstract dataset and 13% of those in Results are cross-sentence relations. We did not consider them in this work, which limits the performance upper bound that can be achieved. Second, while we attempted to normalize entities to standard terminologies (ie, entity linking[57]) in order to also generate a document-level dataset similar to BioRED,[29] a significant proportion of entities (about 32%) could not be mapped to unique identifiers, highlighting the lack of standardization of nutrition and microbiome-related terms in ontologies.[23] We leave the construction of this dataset for future work. Lastly, the end-to-end model performance is modest, partly due to the relatively large number of entity and relation types, we considered and partly due to complex syntactic structures we observed in our dataset.

## Conclusions

We constructed an annotated corpus of diet-microbiome associations in the scientific literature. To our knowledge, this is the largest and most diverse corpus in this domain and serves as a valuable resource for advancing the understanding of the complex interactions between diet, human metabolism, and gut microbiota.

Our experimental results show that state-of-the-art NLP models exhibit promising performance in NER, but that the end-to-end model performance remains suboptimal, highlighting both the challenges with the biomedical RE task generally and with the complex text common in diet-microbiome literature. Future work will involve extending the corpus to document level to make it more suitable for generating knowledge graphs, improving NER and trigger detection models to improve RE performance, implementing entity linking models, and enhancing the LLM-based method to better deal with NULL relations.

## Acknowledgments

We thank Evan Guerra for reproducing the experiments.

## Author contributions

Gibong Hong (Data curation, Formal analysis, Investigation, Methodology, Software, Validation, Visualization, Writing—original draft, Writing—review & editing), Veronica Hindle (Data curation, Writing—review & editing), Nadine M. Veasley (Data curation, Writing—review & editing), Hannah D. Holscher (Conceptualization, Data curation, Funding acquisition, Supervision, Writing—review & editing), and Halil Kilicoglu (Conceptualization, Data curation, Funding acquisition, Methodology, Project administration, Resources, Supervision, Writing—original draft, Writing—review & editing)

## Supplementary material

Supplementary material is available at *Journal of the American Medical Informatics Association* online.

## Funding

This work was supported by a University of Illinois Personalized Nutrition Initiative Seed Grant (H.K. and H.D.H.). It was also partially supported by funding from the National Center for Complementary and Integrative Health (NCCIH) and the Office of Data Science Strategy (ODSS) (grant number: U01AT012871) (H.K.). Its contents are solely the responsibility of the authors and do not necessarily represent the official views of the NCCIH, ODSS, and the National Institutes of Health. This study used Bridges-2 and Ocean at Pittsburgh Supercomputing Center through allocation #CIS230099 from the Advanced Cyberinfrastructure Coordination Ecosystem: Services & Support (ACCESS) program,[58] supported by National Science Foundation grants #2138259, #2138286, #2138307, #2137603, and #2138296.

## Conflicts of interest

The authors state that they have no competing interests to declare.

## Data availability

DiMB-RE and the NLP models are available at https://github.com/ScienceNLP-Lab/DiMB-RE.